\documentclass[11pt]{article}
\title{USING WORDNET TO COMPLEMENT TRAINING INFORMATION IN TEXT
CATEGORIZATION\thanks{This research has been supported by the Spanish
Commitee of Science and Technology (CICYT TIC94-0187) - Published in
{\em Recent Advances in Natural Language Processing}, 1997}}
\author{Manuel de Buenaga Rodr\'{\i}guez, Jos\'e Mar\'{\i}a
G\'omez-Hidalgo,\\ Bel\'en D\'{\i}az-Agudo\\
Departamento de Sistemas Inform\'aticos y Programaci\'on\\
Universidad Complutense de Madrid\\
Avda. Complutense s/n, 28040 Madrid (Spain)\\
\{mbuenaga,jmgomez\}@dia.ucm.es, belend@eucmax.sim.ucm.es}
\date{}
\begin{document}

\maketitle

\begin{abstract}
Automatic Text Categorization (TC) is a complex and useful task for
many natural language applications, and is usually performed through
the use of a set of manually classified documents, a {\em training
collection}.  We suggest the utilization of additional resources like
lexical databases to increase the amount of information that TC
systems make use of, and thus, to improve their performance.  Our
approach integrates {\sc WordNet} information with two training
approaches through the Vector Space Model.  The training approaches we
test are the Rocchio (relevance feedback) and the Widrow-Hoff (machine
learning) algorithms.  Results obtained from evaluation show that the
integration of {\sc WordNet} clearly outperforms training approaches,
and that an integrated technique can effectively address the
classification of low frequency categories.\\
\end{abstract}

\section{Introduction}

{\em Text categorization} (TC) is the classification of documents with
respect to a set of one or more pre-existing categories.  TC is a hard
and very useful operation frequently applied to the assignment of
subject categories to documents, to route and filter texts, or as a
part of natural language processing systems.

Most categorization systems make use of a training collection of
documents to predict the assignment of new documents to categories.
We propose the utilization of additional resources to increase the
amount of information that TC systems make use of, and thus, to
improve their effectiveness.  We have selected the lexical database
{\sc WordNet} to integrate it with the use of a training collection.

In order to test the hypothesis that the utilization of lexical
databases improves a training-based TC system, we have performed a
series of experiments on the Reuters-21578 TC test collection.  Among
many training approaches that have been employed in TC, we have
selected the Rocchio and the Widrow-Hoff algorithms.  We have combined
the utilization of each algorithm with {\sc WordNet}, using the Vector
Space Model for this task.  The results obtained evaluating both
hybrid systems show that:

\begin{itemize}

\item An integrated approach combining a training collection and a
lexical database performs better than the isolated use of a training
collection.

\item A combined approach to TC can effectively address the
classification of documents into low frequency categories, even if few
or none training data is available for these categories.

\end{itemize}

This work is organized as follows.  First of all, we introduce the TC
task and the resources we make use of.  Next, we describe the model in
which these elements are integrated.  After this, we examine both
training approaches and how to integrate {\sc WordNet} into them.
Next we present our evaluation environment and results.  Related work
is later discussed, and finally, we describe our conclusions and lines
of future work.

\section{Task Description}

Given a set of documents and a set of categories, the goal of a
categorization system is to decide whether any document belongs to any
category or not.  The system makes use of the information contained in
a document to compute a degree of pertinence of the document to each
category.  Categories are usually subject labels like {\sc art} or
{\sc military}, but other categories like text genres are also
interesting \cite{karlgren94}.  Documents can be news stories, e-mail
messages, reports, and so forth.

The most widely used resource for TC is the training collection.  A
{\em training collection} is a set of manually classified documents
that allows the system to guess clues on how to classify new unseen
documents.  There are currently several TC test collections, from
which a training subset and a test subset can be obtained.  For
instance, the huge TREC collection \cite{harman96}, OHSUMED
\cite{hersh94} and Reuters-21578 (new release of Reuters-22173
\cite{lewis92}) have been collected for this task.  We have selected
Reuters because it has been used in other work, facilitating the
comparison of results.

Effectiveness of training approaches to TC depends on the number of
training examples per category \cite{larkey96}.  Categories used to
classify training collections can have few training documents.
Training-based TC approaches usually get worse results for these
categories \cite{lewis96,larkey96}.  However, these categories have
been designed by experts to be used in retrieval \cite{lowe94}.  We
propose the utilization of lexical databases for improving results for
all categories and especially those with few training examples.

Lexical databases have been rarely employed in TC, but several
approaches have demonstrated their usefulness for term classification
operations like {\em word sense disambiguation}
\cite{resnik95,agirre96}.  A {\em lexical database} is a reference
system that accumulates information on the lexical items of one or
several languages.  In this view, machine readable dictionaries can
also be regarded as primitive lexical databases.  Current lexical
databases include {\sc WordNet} \cite{miller95}, EDR \cite{yokoi95}
and Roget's Thesaurus.  {\sc WordNet}'s large coverage and frequent
utilization has led us to use it for our experiments.

\section{The Vector Space Model for Text Categorization}

The {\em Vector Space Model} (VSM) \cite{salton83} was originally
developed for Information Retrieval, but it provides support for many
text classification tasks.  In fact, the VSM is a very suitable
environment for expressing our approach to TC \cite{gomez97}.  Also,
it is supported by many experiences in text retrieval
\cite{lewis92,salton89}.

The bulk of the VSM for IR is representing natural language
expressions as term weight vectors.  Each weight measures the
importance of a term in a natural language expression, which can be a
document or a query.  Semantic closeness between documents and queries
is computed by the cosine of the angle between document and query
vectors.  We have noted an analogy between queries in IR and
categories in TC, that allows to easily adapt the VSM to TC.
Categories can be also represented by term weight vectors, and the
cosine formula used to compute the similarity between documents and
categories.

Given three sets of $N$ terms, $M$ documents to be classified, and $L$
categories, the weight vector for document $j$ is $\langle wd_{1j},
wd_{2j}, \ldots , wd_{Nj} \rangle$ and the weight vector for category
$k$ is $\langle wc_{1k}, wc_{2k}, \ldots , wc_{Nk} \rangle$.  The
similarity between document $j$ and category $k$ is obtained with the
formula:

\begin{equation}
sim(d_{j},c_{k}) = \frac{{\displaystyle \sum_{i=1}^{N} wd_{ij}
\cdot wc_{ik}}}{{\displaystyle \sqrt{{\sum_{i=1}^{N} wd_{ij}^{2}}
\cdot {\sum_{i=1}^{N} wc_{ik}^{2}}}}}
\label{sim}
\end{equation}

Nevertheless, like in IR, the VSM does not cover several important
issues in TC: selection of terms for representation, computation of
term weights (both for documents and categories), and definition of an
assignment policy of documents to categories.

\begin{itemize}

\item First, it is possible to select the terms using the term
discrimination model by Salton and others \cite{salton76}, or using
term quality measures like the {\em expected mutual information}
between categories and terms \cite{vanrijsbergen77}.  We have chosen
this latter approach, because it provides terms even for those
categories with less training documents.

\item Secondly, weights for documents vectors can be computed making
use of well known formulae based on term frequencies.  We have used
the expression \cite{salton89}:

\begin{equation}
wd_{ij}=tf_{ij} \cdot tw_{i}
\label{wd}
\end{equation}

Where $tf_{ij}$ is the frequency of term $i$ in document $j$, and
$tw_{i}$ is the weight or importance of the term $i$ in the
collection.  However, all the document to be classified are not always
available in a specific moment of time in real-world problems, so the
weights of terms have to be estimated using an additional resource.
Reckoning of weights for categories vectors also needs the use of an
independent resource like a manually classified document collection or
a lexical database.

\item Finally, simple assignment policies can be defined using the
ranking of documents inside categories.  Our evaluation process does
not depend on the policy selected, as we will see.

\end{itemize}

\section{Using a Training Collection to Represent Categories}

A set of manually classified documents can be used to predict the
assignment of new documents to categories.  Approaches to TC based on
such a training collection include {\em k-nearest-neighbor} algorithms
\cite{masand92}, Bayesian classifiers \cite{lewis92}, neural networks
\cite{wiener95}, and learning algorithms based on relevance feedback
or from the field of machine learning \cite{lewis96}, or based in
decision trees \cite{apte94}.  Many of these approaches can be
employed in the VSM for TC, as we have introduced this model.  Many of
them could also facilitate the integration of an independent resource
like a lexical database in the process of TC.

Training algorithms provide a way to calculate the weights for
categories vectors.  The basic idea is using a training formula that
assigns a weight to a term in a category vector, in proportion to the
number of occurrences of the term in documents manually assigned to the
category, and to the importance of the term in the collection too.

We have selected the Rocchio \cite{rocchio71} and the Widrow-Hoff
\cite{widrow85} algorithms to compute the term weights for categories
in our approach.  The first one is an algorithm traditionally used for
Relevance Feedback in IR. The second one comes from Machine Learning,
and it has been recently used in TC, outperforming the Rocchio one
\cite{lewis96}.  Both algorithms give the chance of integrating an
initial representation computed by the utilization of an external
resource like {\sc WordNet}.  Nevertheless, as far as we are
concerned, an important difference exists between Rocchio and
Widrow-Hoff algorithms.  The former assigns the same importance to
training for each category, even if it has very few training
instances.  The latter, however, produces greater training weight for
categories with many training instances, and lower weights when
training documents are few.  This second approach leads to a more
coherent integration of the training-computed weights with the
independently computed ones.

We show how to calculate the weights for category vectors using both
of these algorithms.  We suppose the existence of a set of $P$
training documents, previously represented using an analogous formula
to (\ref{wd}), the one used for those documents to be classified.  The
weight of the term $i$ in the $l$ document is represented by
$wd_{il}$.

\subsection{The Rocchio Algorithm}

The Rocchio algorithm produces a new weight vector $wc_{k}$ from an
existing one $wc_{k}^{0}$ and a collection of training documents.  The
component $i$ of the vector $wc_{k}$ is computed by the formula:

\begin{eqnarray}
wc_{ik} & = & \alpha wc_{ik}^{0} + \beta \frac{{\sum_{l \in
C_{k}} wd_{il}}}{n_{k}} \nonumber \\
        &   & \mbox{} + \gamma \frac{{\sum_{l \not \in
C_{k}} wd_{il}}}{P - n_{k}}
\label{rocchio}
\end{eqnarray}

Where $wc_{ik}^{0}$ is the initial weight of the term $i$ for the
category $k$, $wd_{il}$ is the weight of the term $i$ for the training
document $l$, $C_{k}$ is the set of indexes of documents assigned to
the category $k$, and $n_{k}$ the number of these documents.  The
parameters $\alpha$, $\beta$ and $\gamma$ control the relative impact
of the initial, positive and negative weights respectively in the new
vector.  As Lewis \cite{lewis96}, we have used the values $\beta = 16$
and $\gamma = 4$.  The value of $\alpha$ is set to $20$, in order to
balance the importance of initial and training weights.  We restrict
the classifier to make no use of negative weights, so the final weight
$wc_{ik}$ will be positive, or turned to 0 if negative.

In TC, the initial vector $wc_{k}^{0}$ is usually a null vector, but
it can be filled with a set of initial weights calculated by the use of
an external resource.  In the next section, we see how to do this
employing WordNet.

\subsection{The Widrow-Hoff Algorithm}

The Widrow-Hoff algorithm starts with an existing weight vector
$wc_{k}^{0}$ and sequentially updates it one time for every training
document.  The component $i$ of the vector $wc_{k}^{l+1}$ is got from
the $lth$ document and from the $lth$ vector by the formula:

\begin{equation}
wc_{ik}^{l+1} = wc_{ik}^{l} + 2 \eta (wd_{l} \cdot wc_{k}^{l} -
y_{l}) wd_{il}
\label{widrowHoff}
\end{equation}

Where $wc_{ik}^{l}$ is the weight of the term $i$ in the $lth$ vector
for category $k$, $wd_{l}$ is the term weight vector for document $l$,
$wc_{k}^{l}$ is the $lth$ vector for category $k$, $y_{l}$ is 1 if the
$lth$ document is assigned to the category $k$ and 0 in other cases,
and $wd_{il}$ is the weight of term $i$ in the $lth$ document.  The
constant $\eta$ is the learning rate, which controls how quickly the
weight vector is allowed to change, and how much influence each new
document has on it.  A value typically used for $\eta$ is
\(1/4X^{2}\), being $X$ the maximum value of the norm of vectors that
represent training documents.

As in Rocchio algorithm, an initial weight vector can be produced
using an independent resource.  However, the importance of this
initial weights is reduced proportionally to the number of training
documents which are available for a category.  When there are many
training examples, this initial weight is dominated by the weight
obtained from these examples.  When there are few training instances,
the initial weights tend to keep their values.

\section{Using {\sc WordNet} to Complement Training Information}

The combination of information from {\sc WordNet} and from the
training collection is performed by the use of initial weights for
categories.  Next we discuss the way we have produced the initial
weights from {\sc WordNet} and how we have integrated them into each
of both algorithms.

\subsection{Obtaining Synonym Information from {\sc WordNet}}

The utilization of {\sc WordNet} is based in the assumption that the
name of a category can be a good predictor of its occurrence.  For
instance, the occurrence the word ``barley'' suggests that a news
article should be classified into the {\sc barley}\footnote{All the
following examples are taken from the Reuters category set and involve
words that actually occur in the documents.} category.  The prediction
of more general categories like {\sc earn} ({\em earnings}) should
instead rely on the occurrence of semantically more independent terms
like ``dollar'' or ``invest.''

Lexical databases contain many kinds of information on lexical items:
concepts; synonymy and other lexical relations; hyponymy and other
conceptual relations; etc.  For instance, {\sc WordNet} represents
concepts as synonyms sets, or {\em synsets}.  Using {\sc WordNet},
synonyms for names of categories can be found, and then used to
predict categories assignments.  A TC system can also exploit lexical
and conceptual relations in {\sc WordNet}, to find terms which are
semantically close to a category.  In an initial approach, we have
focused only on the synonymy relation in {\sc WordNet}.

We have performed a ``category expansion,'' similar to query expansion
in IR. For any category, its closer synsets are selected, and any term
belonging to them is added to the term set.  We have taken only
concepts that are candidates to represent the meaning of each
category, making no use of any conceptual relation in {\sc WordNet}.
The selection of candidate synsets can be considered as a
disambiguation process, and it has been performed manually, because
the small number of categories in our test collection made it
affordable.  We are currently designing algorithms for automating this
operation.

Terms obtained from the selected synsets filtered using a classic
stoplist, and they are stemmed after using the Porter algorithm
\cite{frakes92}.  Those terms that do not occur in any training
document are deleted.  For any term, a degree of semantic closeness to
the category it comes from, is computed through the following
criteria:

\begin{itemize}

\item If the term is a direct synonym of the expression that
represents the category (like the term ``peanut" is a synonym of the
expression {\em groundnut}, which corresponds to the code {\sc
groundnut}), semantic closeness between term and category is set to 1.

\item If the expression that represents a category consists of several
words, the semantic value for any synonym of any of these words is
defined as $1/nc$, being $nc$ the number of words in the expression.
For example, the term ``indicant" is a synonym of the word ``index" in
the expression {\em industrial production index} (corresponding to
category with code {\sc ipi}), and its semantic closeness value is
$1/3$.

\item If several values can be defined between a category and a term,
the greatest one is selected.

\end{itemize}

For the $135$ categories in the Reuters document collection, a set of
$246$ terms has been produced.  Also, we have generated a set of $346$
values of term-category semantic closeness, which have been taken as
an initial representation of categories.  The weight of every term is
calculated making use of the same formula that was used for terms
taken from the training collection.  Thus, if a term was selected from
the training collection, and it is chosen now again, it retains the
same weight.

\subsection{Integrating WordNet Information into Training}

We have combined {\sc WordNet} information with the Rocchio and
Widrow-Hoff algorithms to produce categories representation.  The
values of semantic closeness have been taken as the initial weights
for categories, being these weights refined by the use of training
documents.  To keep the initial weights and the training document
weights the same order, the approach to the integration of {\sc
WordNet} information is different for each algorithm.

Weights for terms in documents are numbers of occurrences multiplied
by weights of terms in the training collection. The weights of terms
in the training collection are computed by the formula:

\begin{equation}
tw_{i} = \log_{2} \frac{P}{tf_{i}}
\label{termweight}
\end{equation}

Where $tf_{i}$ is the number of training documents in which term $i$
occurs.  As in previous equations, $P$ is the number of training
documents.  This is the weight used for any document in our approach,
and thus in formulae (\ref{wd}), (\ref{rocchio}), and
(\ref{widrowHoff}).

For the Rocchio algorithm, we have considered the previously produced
value of semantic closeness as a number of occurrences of a term in a
category, so this value has to be multiplied by the weight of the term
in the collection.  Additionally, since $\alpha=20$ and
$\beta+\gamma=20$, weights for terms in categories are balanced
between {\sc WordNet} and the training collection.

On the other side, the insertion of a term weight for a document in
the Widrow-Hoff algorithm is normalized by the $\eta$ constant.  So,
we have divided the initial weights used for Rocchio among $X$, which
is the maximum value of norms of document vectors.  This technique
keeps again initial and training weights the same order.

\section{Evaluation}

We have chosen a set of very extended metrics and a frequently used
free test collection for our work.  The metrics are {\em recall} and
{\em precision}, and the test collection is, as introduced before,
Reuters-21578.  Before stepping into the actual results, we provide a
closer look to these elements.

\subsection{Evaluation measures}

The kind of rankings produced in the VSM promote {\em recall} and {\em
precision} based evaluation, which is very standardized in IR. Recall
and precision are not so standardized in TC, where most of the
measures used depend on the kind of system that is built (automatic
or semi-automatic classifiers, autonomous systems, etc.).  However,
recall and precision are well known measures and they have been used
before in TC \cite{lewis92,wiener95,larkey96}.  We have computed
precision at 11 recall levels, taking the average precision as the
number which describes the overall performance of each technique.

For precision calculation, we have produced a ranking of documents for
each category, according to their similarity to the category.  Instead
of this technique, a ranking of categories per document can be
generated.  We have used the former because we were interested on
examining separate results for each category.  This approach allows to
split the set of categories into two groups: one that contains
categories with few training examples, and another one which contains
frequent categories.  Precision averages are produced at each recall
level for both sets of categories, and for the complete set of
categories.  So, each category has the same influence in final
results, whether it is very frequent or not.

\subsection{The Reuters-21578 Test Collection}

The Reuters-21578 collection consists of 21,578 newswire articles
collected during 1987 from Reuters.  Documents in Reuters deal with
financial topics, and were classified in several sets of financial
categories by personnel from Reuters Ltd.  and Carnegie Group Inc.
Documents vary in length and number of categories assigned, from 1
line to more than 50, and from none categories to more than 8.  There
are five sets of categories: TOPICS, ORGANIZATIONS, EXCHANGES, PLACES,
and PEOPLE. As others before, we have selected the 135 TOPICS for our
experiments.  An example of news article classified in {\sc bop} ({\em
balance of payments}) and {\sc trade} is shown in
Figure~\ref{reuters1}.  Current version of Reuters is marked up with a
Standard Generalized Markup Language (SGML).  Some spurious formatting
and superfluous marks have been removed from the example.

\begin{figure}
\begin{flushleft}
{\small {\tt <REUTERS TOPICS="YES" LEWISSPLIT="TEST"
CGISPLIT="TRAINING-SET" OLDID="6505" NEWID="18753">

<DATE>18-JUN-1987 11:44:27.20</DATE>

<TOPICS><D>bop</D><D>trade</D></TOPICS>

<PLACES><D>italy</D></PLACES>

<TEXT>

<TITLE>ITALIAN BALANCE OF PAYMENTS IN DEFICIT IN MAY</TITLE>

<BODY>

Italy's overall balance of payments showed a deficit of 3,211 billion
lire in May compared with a surplus of 2,040 billion in April,
provisional Bank of Italy figures show.

The May deficit compares with a surplus of 1,555 billion lire in the
corresponding month of 1986.

For the first five months of 1987, the overall balance of payments
showed a surplus of 299 billion lire against a deficit of 2,854
billion in the corresponding 1986 period.

REUTER

</BODY>

</TEXT>}
}
\end{flushleft}
\caption{Document number 18753 from Reuters-21578.} \label{reuters1}
\end{figure}

When a test collection is provided, it is customary to divide it into
a training subset and a test subset.  Several partitions have been
suggested for Reuters \cite{lewis92}, among which ones we have opted
for the Lewis (LEWISSPLIT) one.  First $13,625$ news stories are used
for training, and last $6,188$ are kept for testing (rest of documents
are not used).  We summarize significative statistics about this split
in Table~\ref{reuters2}.  This $13,625/6,188$ partition has been used
before \cite{lewis92} and involves the general case of documents with
no categories assigned.

\begin{table*}
\begin{center}
\begin{tabular}{l|l|r|r|r} \hline \hline
\multicolumn{2}{c|}{}
  & \multicolumn{3}{c}{\em Subcollection} \\ \cline{3-5}
\multicolumn{2}{c|}{}
  & {\em Training} & {\em Test} & {\em Total} \\ \hline
Docs. & Number & 13,625 & 6,188 & 19,813 \\
Words & Occurrences & 1,820,881 & 746,726 & 2,567,607 \\
 & Doc. average & 133 & 120 & 129 \\
Docs. with 1+ Topics & Number & 7,780 & 3,022 & 10,802 \\
 & Percentage & 57 & 48 & 54 \\
Topics & Occurrences & 9,666 & 3,768 & 13,434 \\
 & Doc. Average & 0.70 & 0.60 & 0.67 \\
\hline\hline
\end{tabular}
\end{center}
\caption{ Reuters-21578 document collection statistics.}
\label{reuters2}
\end{table*}

Categories in the TOPICS set include subject codes like {\sc interest}
({\em interest rates}), economic indicator codes like {\sc ipi} ({\em
Industrial Production Index}), currency codes like {\sc escudo} ({\em
Portuguese Escudo}), corporate codes like {\sc acq} ({\em
mergers/acquisitions}), commodity codes like {\sc silver}, and energy
codes like {\sc propane}.  The number of documents assigned to these
categories in the document collection ranges vastly.  For example, the
frequency of the codes in the training subset ranges from $0$ ({\sc
escudo}) to $2,877$ ({\sc earn}), with an average of $71.6$ documents
per category, but $77$ categories have less than $10$ training
examples.  From the $93$ TOPICS with one or more test examples, $33$
categories have less than $10$ training instances, and $60$ categories
have $10$ or more training documents.  This distinction is
interesting because approaches based on training usually ignore
categories with few training examples \cite{lewis92,lewis96,larkey96}.

\subsection{Results and Interpretation}

Results of our series of experiments are introduced in the
Table~\ref{results1}.  This table shows precision at eleven recall
levels for the four approaches we have tested: the Rocchio and
Widrow-Hoff algorithms and the combination of each one with {\sc
WordNet}.  Precision is calculated for the $93$ categories with one or
more test documents, and then an average is obtained.

\begin{table}
\begin{center}
\begin{tabular}{c|cc|cc} \hline\hline
  & \multicolumn{2}{c|}{\em Train.} &
\multicolumn{2}{c}{\em Train. $+$ {\sc WNet}.} \\ \cline{2-5}
 & {\em Rocch.} & {\em WHoff.} & {\em Rocch.} & {\em
WHoff.} \\ \hline
0.0 & 0.567 & 0.565 & 0.733 & 0.703 \\
0.1 & 0.478 & 0.484 & 0.703 & 0.659 \\
0.2 & 0.423 & 0.427 & 0.661 & 0.610 \\
0.3 & 0.362 & 0.375 & 0.601 & 0.555 \\
0.4 & 0.315 & 0.331 & 0.573 & 0.530 \\
0.5 & 0.270 & 0.279 & 0.556 & 0.511 \\
0.6 & 0.224 & 0.225 & 0.503 & 0.469 \\
0.7 & 0.175 & 0.179 & 0.416 & 0.436 \\
0.8 & 0.147 & 0.149 & 0.359 & 0.412 \\
0.9 & 0.119 & 0.122 & 0.296 & 0.351 \\
1.0 & 0.109 & 0.111 & 0.201 & 0.289 \\ \hline
Avg. & 0.290 & 0.295 & 0.509 & 0.502 \\ \hline\hline
\end{tabular}
\end{center}
\caption{Overall results from our experiments.} \label{results1}
\end{table}

The Table~\ref{results1} shows much better results for approaches
combining resources than for approaches based only on training.  With
the integration of {\sc WordNet}, average precision achieves an
improvement of $20$ points for both algorithms.  However, none of the
training approaches performs definitely better than the other one,
neither isolatedly nor combined with {\sc WordNet}.

Since we have also used categories with few training documents for our
evaluation process, we provide a closer look to the results produced
for them.  In the Table~\ref{results2}, average precision is shown for
each approach we tested, computed separately for categories with less
than $10$ training examples and for categories with $10$ or more
training instances.  General results are also offered.

\begin{table}
\begin{center}
\begin{tabular}{l|c|c|c} \hline\hline
 & {\em $<10$} & {\em $\geq 10$} & {\em Total} \\ \hline
Rocchio & 0.276 & 0.297 & 0.290 \\
Widrow-Hoff & 0.278 & 0.305 & 0.295  \\
Rocchio + WN & 0.417 & 0.560 & 0.509  \\
Widrow-Hoff + WN & 0.482 & 0.514 & 0.502  \\ \hline\hline
\end{tabular}
\end{center}
\caption{Results broken down for categories with few and with more
training documents.} \label{results2}
\end{table}

Precision for categories with few training documents is again better
when using {\sc WordNet} than when using only a training collection.
But, to our view, the greatest achievement of our integrated approach
for low frequency categories is that their results are competitive.
With the utilization of {\sc WordNet}, TC systems can deal better with
all categories proposed for the problem.  However, it should be
pointed out that the behavior of both algorithms seems different.
Widrow-Hoff algorithm shows more uniform results than Rocchio one, a
point that we will study in future work.

\section{Related Work}

Text categorization has emerged as a very active field of research in
the recent years.  Many studies have been conducted to test the
accuracy of training methods, although much less work has been
developed in lexical database methods.  However, lexical databases and
especially {\sc WordNet} have been often used for other text
classification tasks, like word sense disambiguation.

Many different algorithms making use of a training collection have
been used for TC, which have been mentioned in Section 4.  A close
approach to ours is the one from Larkey and Croft \cite{larkey96}, who
combine {\em k-nearest-neighbor}, Bayesian independent and relevance
feedback classifiers, showing improvements over the separated
approaches.  Although they do not make use of several resources, their
approach tends to increase the information available to the system, in
the spirit of our ideas.  Apart from this, Lewis and colleagues have
used Rocchio, Widrow-Hoff and {\em exponentiated-gradient} algorithms
for developing linear classifiers for TC and Text Routing
\cite{lewis96}.  This approach inspired us the utilization of Machine
Learning algorithms, although Lewis' and colleagues' evaluation
techniques and test collections do no allow the comparison of results.

To our knowledge, lexical databases have been used only once before in
TC, apart from our previous work.  Hearst \cite{hearst94} adapted a
disambiguation algorithm by Yarowsky using {\sc WordNet} to recognize
category occurrences.  Categories are made of {\sc WordNet} terms,
which is not the general case of standard or user-defined categories.
It is a hard task to adapt {\sc WordNet} subsets to pre-existing
categories, especially when they are domain dependent.  Hearst's
approach has shown promising results confirmed by our previous work
\cite{gomez97} and present results.

Lexical databases have been employed recently in word sense
disambiguation.  For example, Agirre and Rigau \cite{agirre96} make
use of a semantic distance that takes into account structural factors
in {\sc WordNet} for achieving good results for this task.
Additionally, Resnik \cite{resnik95} combines the use of {\sc WordNet}
and a text collection for a definition of a distance for
disambiguating noun groupings.  Although the text collection is not a
training collection (in the sense of a collection of manually labeled
texts for a pre-defined text processing task), his approach can be
regarded as the most similar to ours in the disambiguation setting.
Finally, Ng and Lee \cite{ng96} make use of several sources of
information inside a training collection (neighborhood, part of
speech, morphological form, etc.)  to get good results in
disambiguating unrestricted text.

All in all, we can see that combining resources in TC is a new and
promising approach supported by previous research in this and other
text classification operations.  We believe that automatic TC
integrating several resources will compete with manual indexing in
quality, and beat it in cost and efficiency.

\section{Conclusions and Future Work}

In this paper, we have presented a new approach to TC based on the
integration of resources to improve the effectiveness.  This approach
integrates the information from the lexical database {\sc WordNet}
into Rocchio and Widrow-Hoff training algorithms through a VSM for TC.
The technique is based on improving the representation of categories
construction through the use of the lexical database, which overcomes
training deficiencies.  We have tested our approach with the
Reuters-21578 TC test collection, achieving two conclusions: first,
combined approach performs much better than those based only in
training; and secondly, with the utilization of lexical databases,
categories with few training documents have no longer to be ignored.

Two main work lines are open: first, we have to conduct new series of
experiments to explain why the integration of {\sc WordNet} into each
training algorithm drives to different results in categories with few
training examples; second, we plan to integrate more {\sc WordNet}
information (like hyperonymy and meronymy relations) with training
approaches and to evaluate approaches based only on {\sc WordNet}.

\end{document}